\documentclass[11pt,a4paper]{article}
\usepackage{times,latexsym}
\usepackage{url}
\usepackage[T1]{fontenc}

\usepackage[acceptedWithA]{tacl2021v1}

\usepackage{graphicx}
\usepackage{caption}
\usepackage{subcaption}
\usepackage{tabularx}
\usepackage{amsmath}
\usepackage{amssymb}
\usepackage{booktabs}
\usepackage{pbox}
\usepackage{multirow}

\usepackage{csquotes}
\usepackage{times}
\usepackage{latexsym}

\usepackage[T1]{fontenc}

\usepackage[utf8]{inputenc}

\usepackage{microtype}

\title{In-Context Retrieval-Augmented Language Models}

\author{Ori Ram\thanks{\;\;Equal contribution.}~~~~~Yoav Levine$^*$~~~~Itay Dalmedigos~~~~Dor Muhlgay\\
\textbf{Amnon Shashua~~~~Kevin Leyton-Brown~~~~Yoav Shoham}\\
  AI21 Labs\\
  \texttt{\{orir,yoavl,itayd,dorm,amnons,kevinlb,yoavs\}@ai21.com} }

\begin{document}
\maketitle

\begin{abstract}

Retrieval-Augmented Language Modeling (RALM) methods, which condition a language model (LM) on relevant documents from a grounding corpus during generation, 
were shown to significantly improve language modeling performance. In addition, they can mitigate the problem of factually inaccurate text generation and provide natural source attribution mechanism.
Existing RALM approaches focus on modifying the LM architecture in order to facilitate the incorporation of external information, significantly complicating deployment.
This paper considers a simple alternative, which we dub \emph{In-Context RALM}: leaving the LM architecture unchanged and prepending grounding documents to the input, \emph{without any further training of the LM}. 
We show that In-Context RALM that builds on off-the-shelf general purpose retrievers provides surprisingly large LM gains across model sizes and diverse corpora. 
We also demonstrate that the document retrieval and ranking mechanism can be specialized to the RALM setting to further boost performance. 
We conclude that In-Context RALM has considerable potential to increase the prevalence of LM grounding, particularly in settings where a pretrained LM must be used without modification or even via API access.\footnote{Our code is available at \url{https://github.com/AI21Labs/in-context-ralm}
}
\end{abstract}
\section{Introduction}\label{sec:1}
\begin{figure}[t]
\centering
\hspace*{-11pt}
\includegraphics[width=1.06\columnwidth]{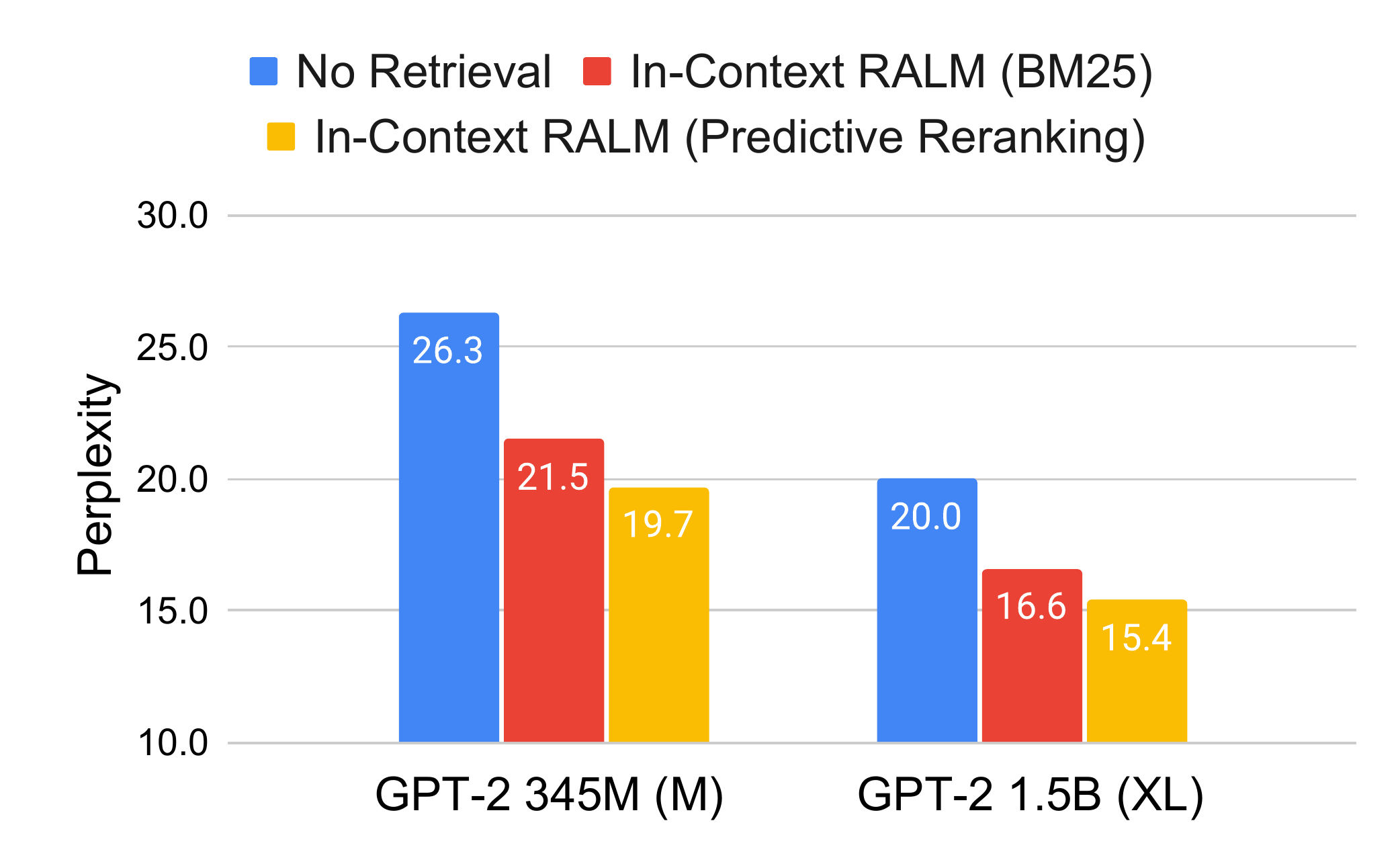}
\vspace{-15pt}
\caption{Our framework, dubbed \emph{In-Context RALM}, provides large language modeling gains on the test set of WikiText-103, \textit{without modifying the LM}.
Adapting the use of a BM25 retriever~\cite{bm25} to the LM task (\S\ref{sec:5}) yields significant gains, and choosing the grounding documents via our new class of Predictive Rerankers (\S\ref{sec:6}) provides a further boost.
 See Table~\ref{tab:retrieval_results} for the full results on five diverse corpora.
}
\label{fig:intro_results}
\end{figure}
\begin{figure*}[t!]
\centering
\includegraphics[width=\textwidth]{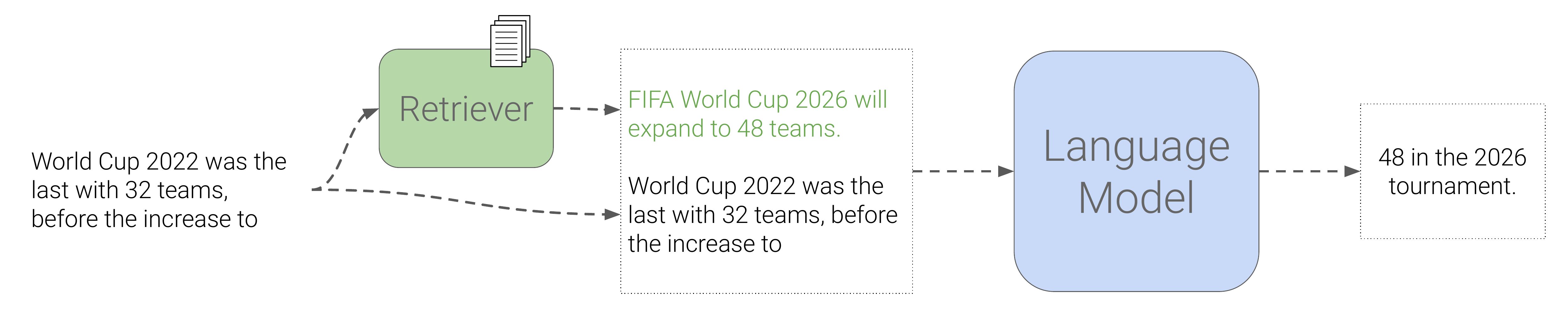}
\vspace{-20pt}
\caption{\small An example of \textit{In-Context RALM}: we simply prepend the retrieved document before the input prefix. }
\label{fig:icralm_example}
\end{figure*}

Recent advances in language modeling (LM) 
have dramatically increased the usefulness of machine-generated text across a wide range of use-cases and domains \cite{gpt3}. 
However, the mainstream paradigm of generating text with LMs bears  
inherent limitations in access to external knowledge.
First, 
LMs are not coupled with any source attribution, and must be trained in order to incorporate up-to-date information that was not seen during training.
More importantly, they tend to produce factual inaccuracies and errors \cite{lin-etal-2022-truthfulqa,maynez-etal-2020-faithfulness,huang2020challenges}.
This problem is present in any LM generation scenario, and is exacerbated when generation is made in uncommon domains or private data.
A promising approach for addressing the above is Retrieval-Augmented Language Modeling (RALM), grounding the LM during generation by 
conditioning on relevant documents retrieved from an external knowledge source. 
RALM systems include two high level components: (i) \emph{document selection}, selecting the set of documents upon which to  condition; and (ii) \emph{document reading}, determining how to incorporate the selected documents into the LM generation process.

Leading RALM systems introduced recently tend to be focused on altering the language model architecture~\cite{knn-lm,retro,zhong-etal-2022-training,levine2021inductive,li2022decoupled}. 
Notably,~\citet{retro} introduced RETRO, featuring document reading via nontrivial modifications that require further training to the LM architecture, while using an off-the-shelf frozen BERT retriever for document selection. Although the paper's experimental findings showed impressive performance gains, 
the need for changes in architecture and dedicated retraining has hindered the wide adoption of such models.

In this paper, we show that a very simple document reading mechanism can have a large impact, and that substantial gains can also be made by adapting the document selection mechanism to the task of language modeling. Thus, we show that many of the benefits of RALM can be achieved while working with off-the-shelf LMs, even via API access. 
Specifically, we consider a simple but powerful RALM framework, dubbed \emph{In-Context RALM} (presented in Section \ref{sec:3}), which employs a zero-effort document reading mechanism: we simply prepend the selected documents to the LM’s input text (Figure~\ref{fig:icralm_example}).

Section \ref{sec:4} describes our experimental setup. To show the wide applicability of our framework, we performed LM experiments on a suite of five diverse corpora: {WikiText-103} \cite{wikitext}, {RealNews} \cite{realnews}, and three datasets from The Pile \cite{pile}: {ArXiv}, {Stack Exchange} and {FreeLaw}. We use open-source LMs ranging from  110M to 66B parameters (from the GPT-2, GPT-Neo, OPT and LLaMA model families). 

In Section \ref{sec:5} we evaluate the application of off-the-shelf retrievers to our framework. In this minimal-effort setting, we found that In-Context RALM led to LM performance gains equivalent to increasing the LM's number of parameters by $2$--$3\times$ across all of the text corpora we examined.
In Section~\ref{sec:6} we investigate methods for adapting document ranking to the LM task, a relatively under-explored RALM degree of freedom. Our adaptation methods range from
using a small LM 
to perform zero-shot ranking of the retrieved documents, 
up to training a dedicated bidirectional reranker by employing \textit{self-supervision from the LM signal}. 
These methods lead to further gains in the LM task corresponding to an additional size increase of $2\times$ in the LM architecture.
As a concrete example of the gains, a 345M parameter GPT-2 enhanced by In-Context RALM outperforms a 762M parameter GPT-2 when employing an off-the-shelf BM25 retriever~\cite{bm25}, and outperforms a 1.5B parameter GPT-2 when employing our trained LM-oriented reranker  (see Figure~\ref{fig:intro_results}).
 For large model sizes, our method is even more effective: {In-Context RALM with an off-the-shelf retriever 
 improved the performance of a $6.7$B parameter OPT model to match that of a $66$B parameter parameter OPT model} (see Figure~\ref{fig:opt_results}).

 In Section~\ref{sec:7} we demonstrate the applicability of In-Context RALM to  downstream open-domain questions answering (ODQA) tasks.

In a concurrent work, \citet{shi2023replug} also suggest to augment off-the-shelf LMs with retrieved texts by prepending them to the input. 
Their results are based on training a dedicated retriever for language modeling. 
In contrast, we focus on the gains achievable in using off-the-shelf retrievers for this task.  
We show strong gains of this simpler setting by investigating: (1) which off-the-shelf retriever is best suited for language modeling, (2) the frequency of retrieval operations, and (3) the optimal query length. In addition, we boost the off-the-shelf retrieval performance by introducing two reranking methods that demonstrate further gains in perplexity.

We believe that In-Context RALM can play two important roles in making RALM systems more powerful and more prevalent. 
First, given its simple reading mechanism, In-Context RALM can serve as a clean probe for developing document retrieval methods that are specialized for the LM task. These in turn can be used to improve both In-Context RALM and other more elaborate RALM methods that currently leverage general purpose retrievers. 
Second, due to its compatibility with off-the-shelf LMs, In-Context RALM can help drive wider deployment of RALM systems. 

\section{Related Work}\label{sec:2}

RALM approaches can be roughly divided into two families of models: (i) \textit{nearest-neighbor language models} (also called $k$NN-LM), and (ii) \textit{retrieve and read models}.
Our work belongs to the second family, but is distinct in that it involves no further training of the LM.

\paragraph{Nearest Neighbor Language Models} The $k$NN-LM approach was first introduced in \citet{knn-lm}. The authors suggest a simple inference-time model that interpolates between two next-token distributions: one induced by the LM itself, and one induced by the $k$ neighbors from the retrieval corpus that are closest to the query token in the LM embedding space. \citet{zhong-etal-2022-training} suggest a framework for training these models. While they showed significant gains from $k$NN-LM, the approach requires storing the representations \emph{for each token in the corpus}, an expensive requirement even for a small corpus like Wikipedia. Although numerous approaches have been suggested for alleviating this issue \cite{he-etal-2021-efficient,retomaton}, scaling any of them to large corpora remains an open challenge.

\paragraph{Retrieve and Read Models} This family of RALMs creates a clear division between \textit{document selection} and \textit{document reading} components. 
All prior work involves training the LM.
We begin by describing works that use this approach for tackling downstream tasks, and then mention works oriented towards RALM. 
\citet{rag} and \citet{izacard-grave-2021-leveraging} fine tuned encoder--decoder architectures for downstream knowledge-intensive tasks. \citet{izacard2022atlas} explored different ways of pretraining such models, while~\citet{levine2021inductive} pretrained an autoregressive LM on clusters of nearest neighbors in sentence embedding space. 
\citet{levine2022standing} showed competitive open domain question-answering performance by prompt-tuning a frozen LM as a reader. 
\citet{realm} pretrained REALM, a retrieval augmented \textit{bidirectional, masked} LM, later fine-tuned for open-domain question answering.
The work closest to this paper---with a focus on the language modeling task---is RETRO \cite{retro}, which modifies an autoregressive LM to attend to relevant documents via chunked cross-attention, thus introducing new parameters to the model. 
Our In-Context RALM differs from prior work in this family of models in two key aspects: 
\begin{itemize}
    \item We use \textit{off-the-shelf} LMs for document reading \textit{without any further training of the LM}. 
    \item We focus on \textit{how to choose documents for improved LM performance}. 
\end{itemize}

\section{Our Framework}\label{sec:3}

\subsection{In-Context RALM}\label{sec:3.1}

Language models define probability  distributions over sequences of tokens. Given such a sequence $x_1,...,x_n$, the standard way to model its probability is via next-token prediction:
 $p(x_1,...,x_n) = \prod_{i=1}^n p(x_i|x_{<i})$,
where $x_{<i}:=x_1,...,x_{i-1}$ is the sequence of tokens preceding $x_i$, also referred to as its \emph{prefix}. This autoregressive model is usually implemented via a learned  transformer network~\cite{transformer} parameterized by the set of parameters $\theta$: 
\begin{equation}\label{eq:lm}
 p(x_1,...,x_n) = \prod_{i=1}^n p_{\theta}(x_i|x_{<i}),
\end{equation}
where the conditional probabilities are modeled by employing a causal self-attention mask~\cite{radford2018improving}. Notably, leading LMs such as GPT-2~\cite{gpt2}, GPT-3~\cite{gpt3}, OPT~\cite{zhang2022opt} or Jurassic-1~\cite{lieber2021jurassic} follow this simple parameterization. 

Retrieval augmented language models (RALMs) add an operation that retrieves one or more documents from an external corpus $\mathcal{C}$, and condition the above LM predictions 
on these documents.
Specifically, 
for predicting $x_{i}$, the retrieval operation from $\mathcal{C}$ depends on its prefix: $\mathcal{R}_\mathcal{C}(x_{<i})$, so the most general RALM decomposition is:
 $p(x_1,...,x_n) = \prod_{i=1}^n p(x_i|x_{<i},\mathcal{R}_\mathcal{C}(x_{<i}))$.
 In order to condition the LM generation on the retrieved document, previous RALM approaches used specialized architectures or algorithms (see \S\ref{sec:2}).
Inspired by the success of In-Context Learning \cite{gpt3,dong2023survey}, \textit{In-Context RALM} refers to the following specific, simple method of concatenating the retrieved documents\footnote{We always use a \textit{single document}, but it is conceptually simple to support multiple documents as well.} within the Transformer's input prior to the prefix (see Figure~\ref{fig:icralm_example}), \textit{which does not involve altering the LM weights $\theta$}:   \begin{equation}\label{eq:general_ralm}
\begin{split}
 &p(x_1,...,x_n) = \\
 &\quad\prod_{i=1}^n p_{\theta}\left(x_i|\left[\mathcal{R}_\mathcal{C}(x_{<i});x_{<i}\right]\right),
 \end{split}
 \end{equation}
where $\left[a;b\right]$ denotes the concatenation of strings $a$ and $b$.

Since common Transformer-based LM implementations support limited length input sequences, when the concatenation of the document and the input sequence exceed this limit we remove tokens from the beginning of $x$ until the overall input length equals that allowed by the model. 
Because our retrieved documents are passages of limited length, we always have enough context left from $x$ (see \S\ref{sec:4.3}).

\subsection{RALM Design Choices}\label{sec:3.2}
We detail below two practical design choices often made in RALM systems. In \S\ref{sec:5}, we investigate the effect of these in the setting of In-Context RALM. 
\paragraph{Retrieval Stride} While in the above formulation a retrieval operation can occur at each generation step, we might want to perform retrieval only once every $s>1$ tokens due to the cost of calling the retriever, and the need to replace the documents in the LM prefix during generation. 
We refer to $s$ as the \emph{retrieval stride}. This gives rise to the following In-Context RALM formulation (which reduces back to Eq.~\eqref{eq:general_ralm} for $s=1$):
\begin{equation}\label{eq:ralm}
\begin{split}
    &p(x_1,...,x_n) = \\ &\quad\prod_{j=0}^{n_s-1}\prod_{i=1}^{s} p_{\theta}\left(x_{s\cdot j+i} | \left[\mathcal{R}_\mathcal{C}(x_{\leq s\cdot j});x_{<(s\cdot j+i)}\right]\right),
\end{split}
\end{equation}
where $n_s=n/s$ is the number of retrieval strides.

Notably, in this framework the runtime costs of each retrieval operation is composed of (a) applying the retriever itself, and (b) recomputing the embeddings of the prefix. In \S\ref{sec:5.2} we show that using smaller retrieval strides, \textit{i.e.}, retrieving as often as possible, is superior to using larger ones (though In-Context RALM with larger strides already provides large gains over vanilla LM). Thus, choosing the retrieval stride is ultimately a tradeoff between runtime and performance.

\paragraph{Retrieval Query Length} While the retrieval query above in principle depends on all prefix tokens $x_{\leq s \cdot j}$, the information at the very end of the prefix is typically the most relevant to the generated tokens. If the retrieval query is too long then this information can be diluted. To avoid this, we restrict the retrieval query at stride $j$ to the last $\ell$ tokens of the prefix, \textit{i.e.}, we use $q^{s,\ell}_j:=x_{s \cdot j-\ell+1},...,x_{s \cdot j}$. We refer to $\ell$ as the \emph{retrieval query length}.
Note that prior RALM work couples the retrieval stride $s$ and the retrieval query length $\ell$~\cite{retro}.
In \S\ref{sec:5}, we show that enforcing $s=\ell$ degrades LM performance. 
Integrating these hyper-parameters into the In-Context RALM formulation gives
\begin{equation}\label{eq:ralm-hps}
\begin{split}
    &p(x_1,...,x_n) = \\ &\quad\prod_{j=0}^{n_s-1}\prod_{i=1}^{s} p_{\theta}\left(x_{s\cdot j+i} |  \left[\mathcal{R}_\mathcal{C}(q^{s,\ell}_j);x_{<(s\cdot j+i)}\right]\right).
\end{split}
\end{equation}

\section{Experimental Details}\label{sec:4}

We now describe our experimental setup, including all models we use and their implementation details.

\subsection{Datasets}\label{sec:4.1}
We evaluated the effectiveness of In-Context RALM across five diverse language modeling datasets and two common open-domain question answering datasets. 
\paragraph{Language Modeling} The first LM dataset is \textit{WikiText-103} \cite{wikitext}, which has been extensively used to evaluate RALMs \cite{knn-lm,he-etal-2021-efficient,retro,retomaton,zhong-etal-2022-training}. Second, we chose three datasets spanning diverse subjects from The Pile \cite{pile}: \textit{ArXiv}, \textit{Stack Exchange} and \textit{FreeLaw}. Finally, we also investigated \textit{RealNews} \cite{realnews}, since The Pile lacks a corpus focused only on news (which is by nature a knowledge-intensive domain).

\paragraph{Open-Domain Question Answering}

In order to evaluate In-Context RALM on downstream tasks as well, we use the \textit{Natural Questions} (NQ; \citealt{kwiatkowski-etal-2019-natural}) and \textit{TriviaQA} \cite{joshi-etal-2017-triviaqa} open-domain question answering datasets.

\subsection{Models}\label{sec:4.2}

\paragraph{Language Models} We performed our experiments using the four models of GPT-2 (110M--1.5B; \citealt{gpt2}), three models of GPT-Neo and GPT-J (1.3B--6B; \citealt{gpt-neo,gpt-j}), eight models of OPT (125M--66B; \citealt{zhang2022opt}) and three models of LLaMA (7B--33B; \citealt{touvron2023llama}).
All models are open source and publicly available.\footnote{All models are available for use use via \url{https://huggingface.co/}}

We elected to study these particular models for the following reasons.
The first four (GPT-2) models were trained on WebText \cite{gpt2}, with Wikipedia documents excluded from their training datasets. We were thus able to evaluate our method's ``zero-shot'' performance when retrieving from a novel corpus (for WikiText-103). 
The rest of the models brought two further benefits. First, they allowed us to investigate how our methods scale to models larger than GPT-2. 
Second, the fact that Wikipedia was part of their training data allowed us to investigate the usefulness of In-Context RALM for corpora seen during training.
The helpfulness of such retrieval has been demonstrated for previous RALM methods \cite{knn-lm} and has also been justified theoretically by \citet{levine2021inductive}.

We ran all models with a maximum sequence length of 1,024, even though GPT-Neo, OPT and LLaMA models support a sequence length of 2,048.\footnote{In preliminary experiments, we observed similar improvements from In-Context RALM when using a sequence length of 2,048. We used a sequence length of 1,024 in order to facilitate a direct comparison between all models.}

\paragraph{Retrievers}

We experimented with both sparse (word-based) and dense (neural) retrievers. We used BM25 \cite{bm25} as our sparse model. For dense models, we experimented with (i) a frozen BERT-base \cite{devlin-etal-2019-bert} followed by mean pooling, similar to \citet{retro}; and (ii) the Contriever  \cite{izacard2022unsupervised} and Spider \cite{ram-etal-2022-learning} models, which are dense retrievers that were trained in unsupervised manners. 

\paragraph{Reranking} When training rerankers (Section~\ref{sec:6.2}), we initialized from RoBERTa-base \cite{liu2019roberta}.

\begin{figure}[t]
\centering
\hspace*{-17pt}
\includegraphics[width=1.08\columnwidth]{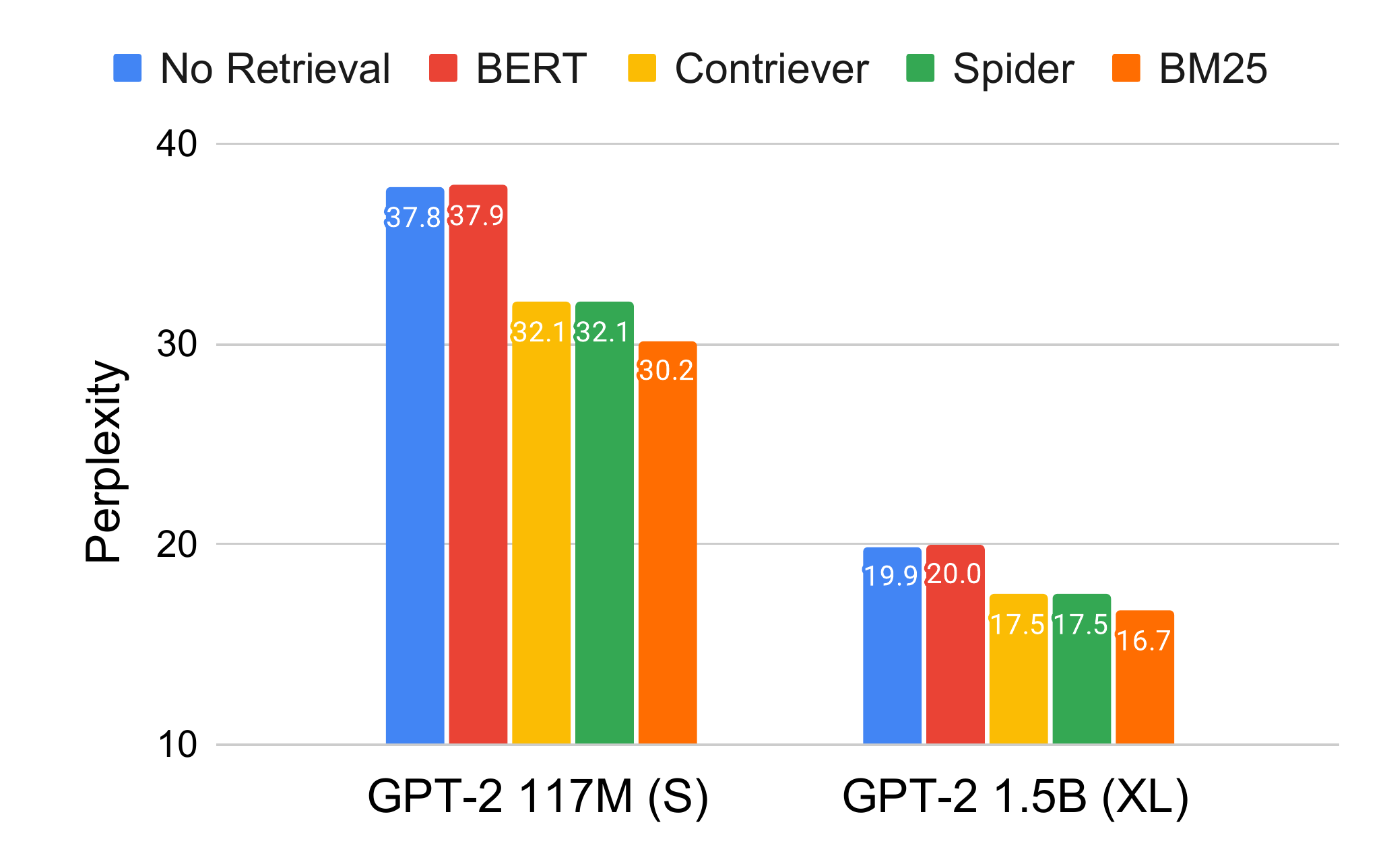}
\vspace{-15pt}
\caption{The performance of four off-the-shelf retrievers used for In-Context RALM on the development set of WikiText-103. All RALMs are run with $s=4$ (\textit{i.e.}, retrieval is applied every four tokens). For each RALM, we report the result of the best query length $\ell$ (see Figures~\ref{fig:query_window},~\ref{fig:bert_query_window},~\ref{fig:contriever_query_window}).}
\label{fig:retrieval-comparison}
\end{figure}
\begin{table*}[t]
\centering
\small
\begin{tabular}{@{}lllccccc@{}}
\toprule
\multirow{2.6}{0pt}{\textbf{Model}} & \multirow{2.6}{35pt}{\textbf{Retrieval}} & \multirow{2.6}{40pt}{\textbf{Reranking}} &
\textbf{WikiText-103} & \textbf{RealNews} & \textbf{ArXiv} & \textbf{Stack Exch.} & \textbf{FreeLaw}  \\
 \cmidrule(lr){4-4} \cmidrule(lr){5-5} \cmidrule(lr){6-6} \cmidrule(lr){7-7} \cmidrule(lr){8-8}
& & & word ppl & token ppl & token ppl & token ppl & token ppl \\
\midrule
\multirow{4.3}{45pt}{\textbf{GPT-2 S}} &
-- & -- & 37.5 & 21.3 & 12.0 & 12.8 & 13.0 \\
& BM25 \S\ref{sec:5} & -- & 29.6 & 16.1 & 10.9 & 11.3 & ~~9.6 \\
& BM25 & Zero-shot \S\ref{sec:6.1} & 28.6 & 15.5  & 10.1 & 10.6 & ~~8.8
\\
& BM25 & Predictive \S\ref{sec:6.2} & 26.8 & ~~~~--  & ~~~~--  & ~~~~-- & ~~~~-- 
\\
\midrule
\multirow{4.3}{45pt}{\textbf{GPT-2 M}} &
-- & -- & 26.3 & 15.7 & ~~9.3 & ~~8.8 & ~~9.6 \\
& BM25 \S\ref{sec:5} & -- & 21.5 & 12.4 & ~~8.6 & ~~8.1 & ~~7.4 \\
& BM25 & Zero-shot \S\ref{sec:6.1} & 20.8 & 12.0  & ~~8.0 & ~~7.7 & ~~6.9
\\
& BM25 & Predictive \S\ref{sec:6.2} & 19.7 & ~~~~--  & ~~~~--  & ~~~~-- & ~~~~-- 
\\
\midrule
\multirow{4.3}{45pt}{\textbf{GPT-2 L}} &
-- & -- & 22.0 & 13.6 & ~~8.4 & ~~8.5 & ~~8.7 \\
& BM25 \S\ref{sec:5} & -- & 18.1 & 10.9 & ~~7.8 & ~~7.8 & ~~6.8 \\
& BM25 & Zero-shot \S\ref{sec:6.1} & 17.6 & 10.6  & ~~7.3 & ~~7.4 & ~~6.4
\\
& BM25 & Predictive \S\ref{sec:6.2} & 16.6 & ~~~~--  & ~~~~--  & ~~~~-- & ~~~~-- 
\\
\midrule
\multirow{4.3}{45pt}{\textbf{GPT-2 XL}} &
-- & -- & 20.0 & 12.4 & ~~7.8 & ~~8.0 & ~~8.0 \\
& BM25 \S\ref{sec:5} & -- & 16.6 & 10.1 & ~~7.2 & ~~7.4 & ~~6.4 \\
& BM25 & Zero-shot \S\ref{sec:6.1} & 16.1 & ~~9.8  & ~~6.8  & ~~7.1 & ~~6.0
\\
& BM25 & Predictive \S\ref{sec:6.2} & 15.4 & ~~~~--  & ~~~~--  & ~~~~-- & ~~~~--  
\\
\bottomrule
\end{tabular}
\caption{
Perplexity on the test set of WikiText-103, RealNews and three datasets from the Pile. For each LM, we report: (a) its performance without retrieval, (b) its performance when fed the top-scored passage by BM25 (\S\ref{sec:5}), and (c) its performance when applied on the top-scored passage of each of our two suggested rerankers (\S\ref{sec:6}). All models share the same vocabulary, thus token-level perplexity (\textit{token ppl}) numbers are comparable. For WikiText we follow prior work and  report word-level perplexity (\textit{word ppl}). 
}
\label{tab:retrieval_results}
\end{table*}

\subsection{Implementation Details}\label{sec:4.3}

We implemented our code base using the Transformers library \cite{wolf-etal-2020-transformers}. We based our dense retrieval code on the DPR repository \cite{karpukhin-etal-2020-dense}.

\paragraph{Retrieval Corpora} For WikiText-103 and ODQA datasets, we used the Wikipedia corpus from Dec. 20, 2018, standardized by \citet{karpukhin-etal-2020-dense} using the preprocessing from \citet{chen-etal-2017-reading}. To avoid contamination, we found and removed all 120 articles of the development and test set of WikiText-103 from the corpus. For the remaining datasets, we used their training data as the retrieval corpus. 
Similar to \citet{karpukhin-etal-2020-dense}, our retrieval corpora consist of non-overlapping passages of 100 words (which translate to less than 150 tokens for the vast majority of passages). Thus, we truncate our retrieved passages at 256 tokens when input to the models, but they are usually much smaller.

\paragraph{Retrieval} For sparse retrieval, we used the Pyserini library \cite{pyserini}. For dense retrieval, we applied exact search using FAISS \cite{faiss2021}.
\begin{figure*}[t!]
\centering
\subfloat{%
  \includegraphics[clip,width=\textwidth]{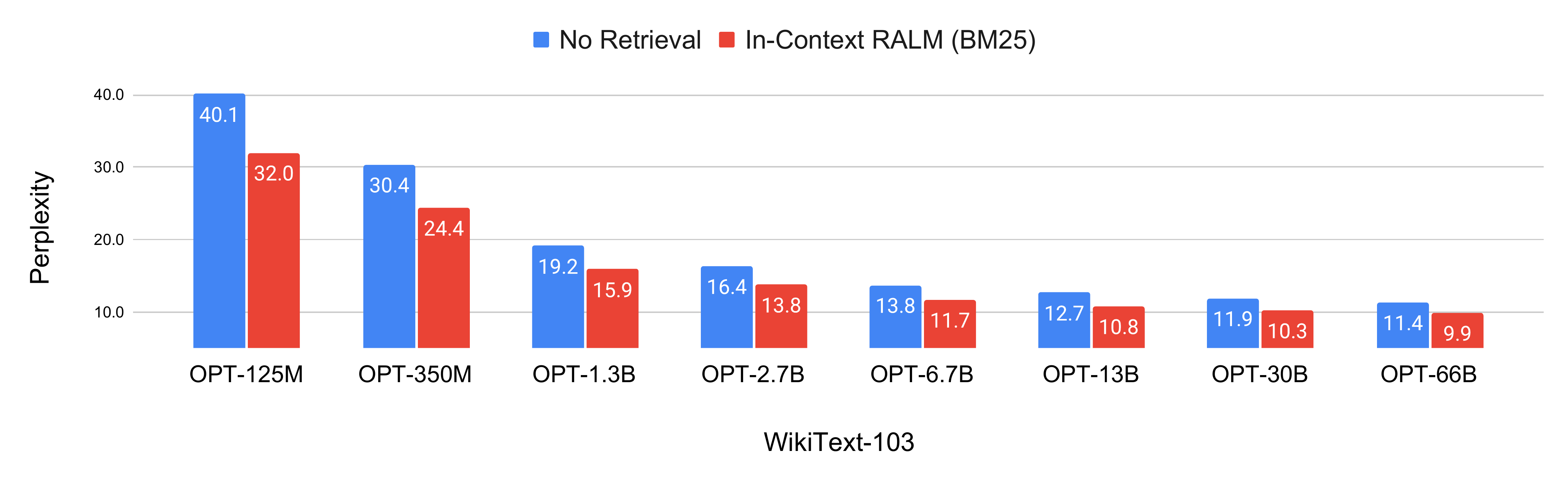}%
}

\subfloat{%
  \includegraphics[clip,width=\textwidth]{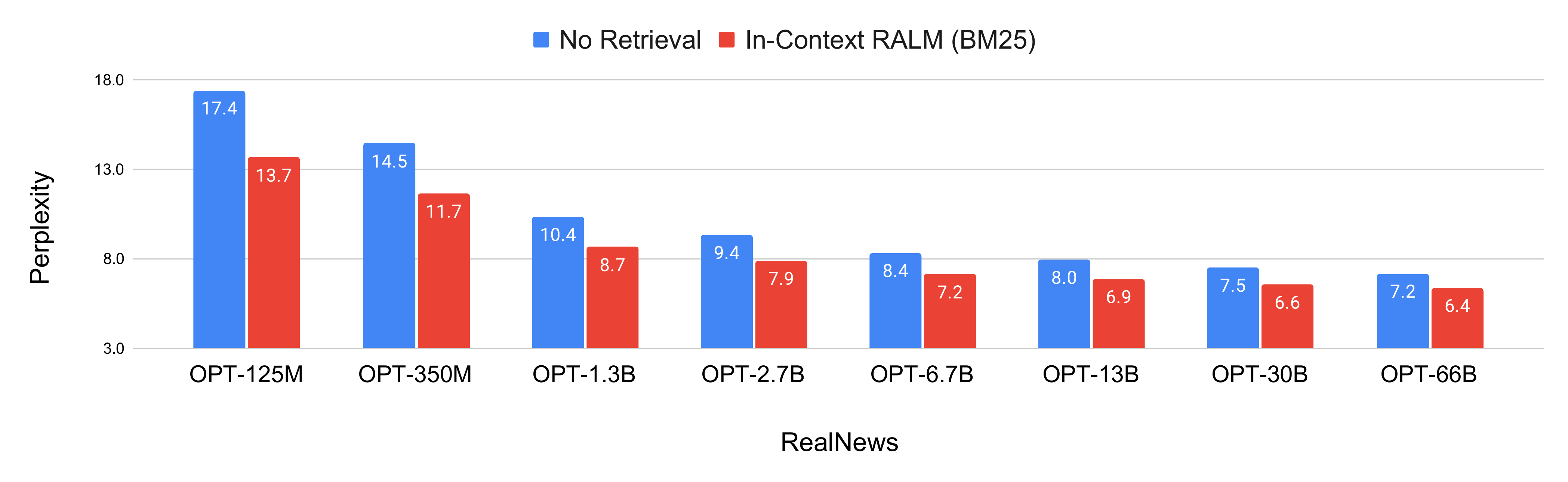}%
}
\caption{Results of OPT models \cite{zhang2022opt} on the test set of WikiText-103 (word-level perplexity) and the development set of RealNews (token-level perplexity).
In-Context RALM models use a BM25 retriever with $s=4$ (\textit{i.e.}, the retriever is called every four tokens) and $\ell=32$ (\textit{i.e.}, the retriever query is comprised of the last 32 tokens of the prefix). \textit{In-Context RALM with an off-the-shelf retriever 
 improved the performance of a $6.7$B parameter OPT model to match that of a $66$B parameter OPT model.} 
}
\label{fig:opt_results}
\end{figure*}

\begin{table}[t!]
\small
\centering
\begin{tabular}{llc}
\toprule
\multirow{2.6}{0pt}{\textbf{Model}} & \multirow{2.6}{33pt}{\textbf{Retrieval}} & 
\textbf{WikiText-103} \\
\cmidrule(lr){3-3} 
& & word ppl \\
\midrule
\multirow{2.3}{63pt}{\textbf{LLaMA-7B} } & - & 9.9 \\
& BM25, \S\ref{sec:5}  & 8.8 \\
\midrule
\multirow{2.3}{60pt}{\textbf{LLaMA-13B}} & - & 8.5 \\
& BM25, \S\ref{sec:5} &  7.6 \\
\midrule
\multirow{2.3}{63pt}{\textbf{LLaMA-33B}} & - & 6.3 \\
& BM25, \S\ref{sec:5}  & 6.1 \\
\bottomrule
\end{tabular}
\caption{The performance of models from the LLaMA family, measured by word-level perplexity on the test set of WikiText-103.
}
\label{tab:results-llama}
\end{table}

\begin{figure}[t]
\centering
\hspace*{-10pt}
\includegraphics[width=1.05\columnwidth]{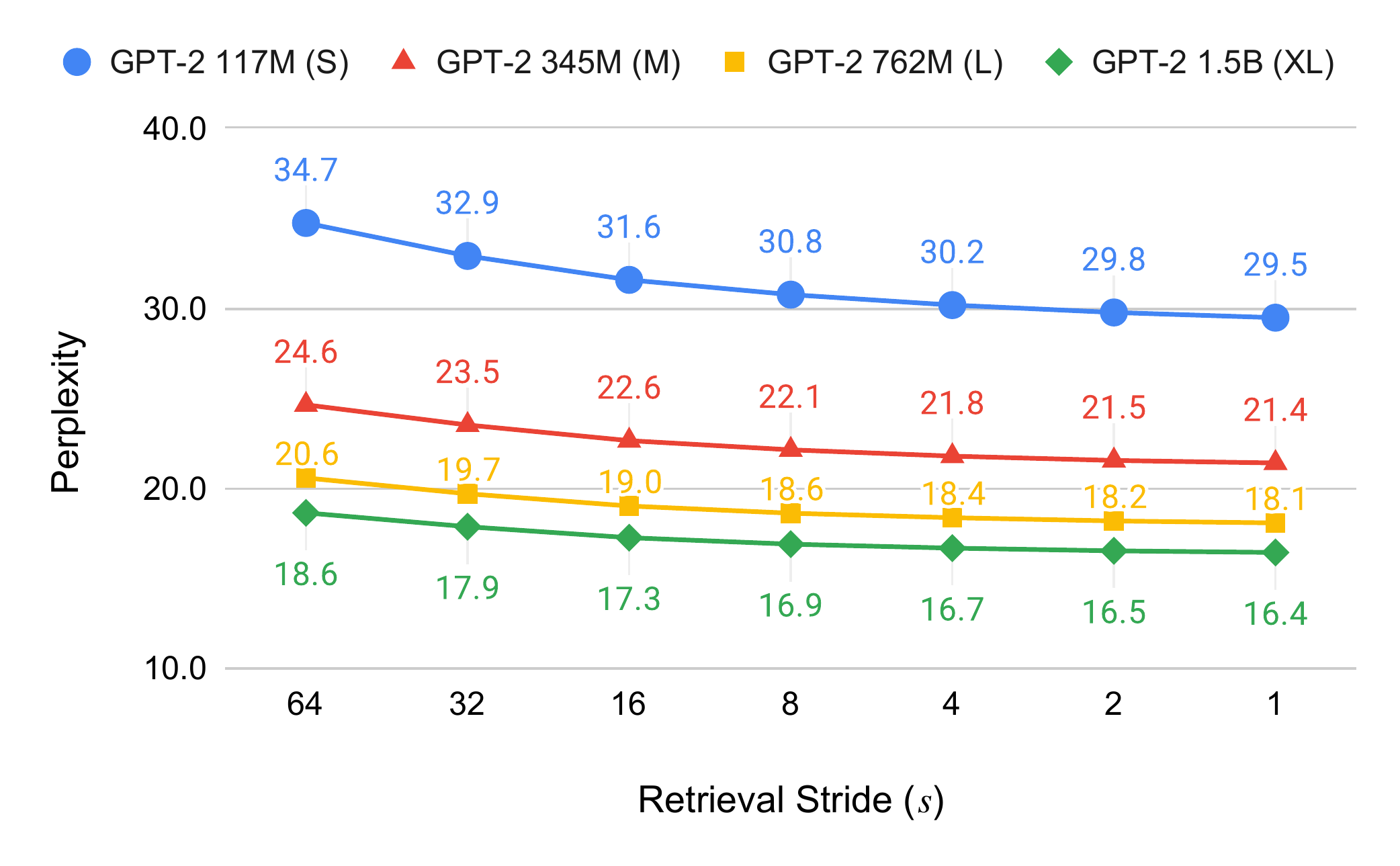}
\vspace{-15pt}
\caption{An analysis of perplexity as a function of $s$, the \emph{retrieval stride}, \textit{i.e.}, the number of tokens between consecutive retrieval operations, on the development set of WikiText-103. Throughout the paper, we use $s=4$ to balance perplexity and runtime.
}
\label{fig:stride}
\end{figure}
\section{The Effectiveness of In-Context RALM with Off-the-Shelf Retrievers}\label{sec:5}

We now empirically show that despite its simple document reading mechanism, In-Context RALM leads to substantial LM gains across our diverse evaluation suite. 
We begin in this section by investigating the effectiveness of off-the-shelf retrievers for In-Context RALM; we go on in \S\ref{sec:6} to show that further LM gains can be made by tailoring document ranking functions to the LM task.

The experiments in this section provided us with a recommended configuration for applying In-Context RALM: applying a sparse BM25 retriever that receives $\ell=32$ query tokens and is applied as frequently as possible. Practically, we retrieve every $s=4$ tokens ($\ell$ and $s$ are defined in \S\ref{sec:3}).  
Table~\ref{tab:retrieval_results} shows for the GPT-2 models that across all the examined corpora, employing In-Context RALM with an off-the-shelf retriever improved LM perplexity to a sufficient extent that it matched that of a $2$--$3\times$ larger model. 
Figure~\ref{fig:opt_results} and Tables~\ref{tab:results-llama} and \ref{tab:results-gpt-neo} show that this trend holds across model sizes up to 66B parameters, for both WikiText-103 and RealNews.

\begin{figure}[t]
\centering
\hspace*{-10pt}
\includegraphics[width=1.05\columnwidth]{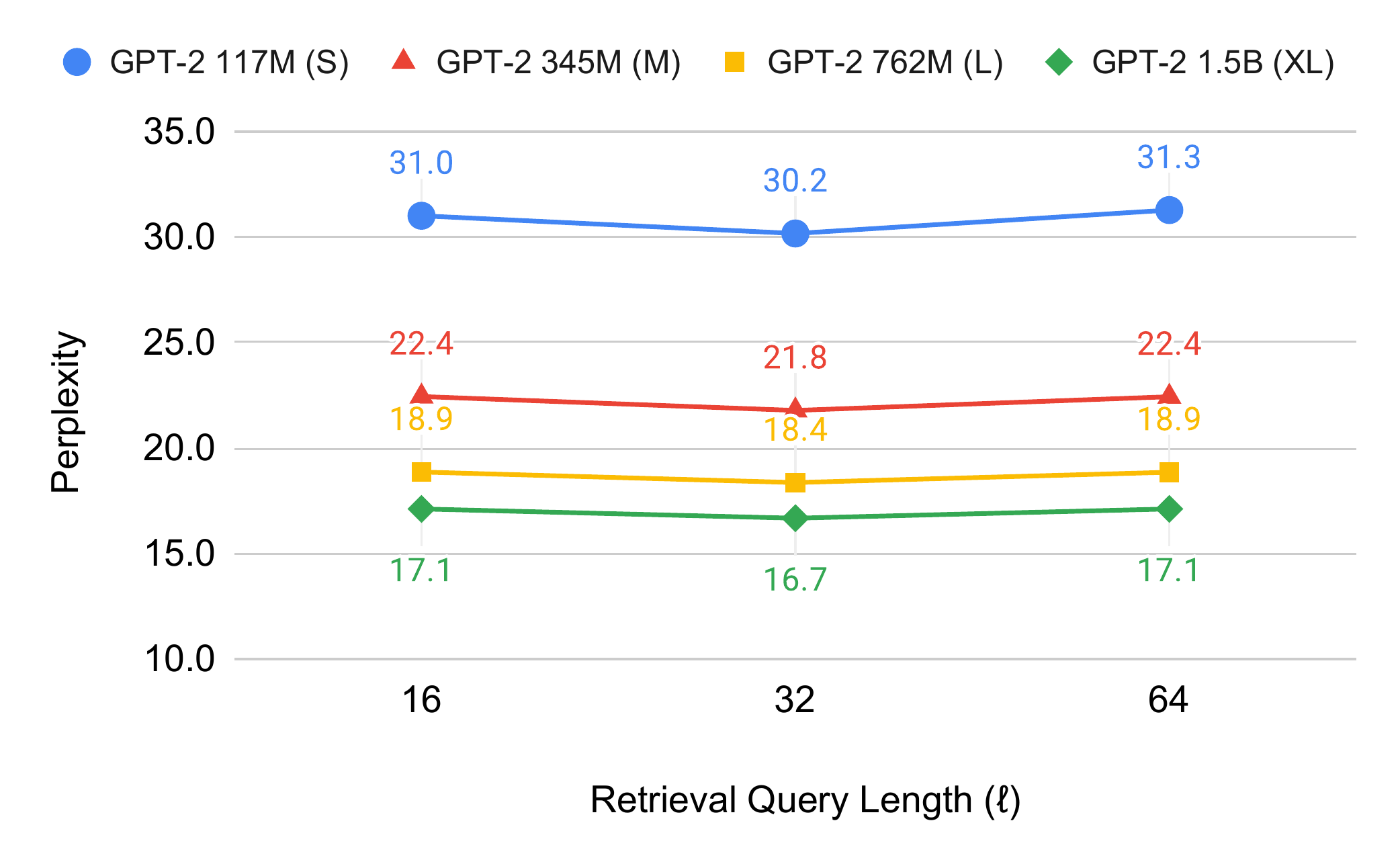}
\vspace{-15pt}
\caption{An analysis of perplexity as a function of \emph{the number of tokens in the query} $\ell$ for BM25  on the development set of WikiText-103.
In the appendix, we show similar trade-offs for dense retrievers within WikiText-103.
Throughout the paper, we use a query length of $\ell=32$ tokens.
}
\label{fig:query_window}
\end{figure}
\subsection{BM25 Outperforms Off-the-Shelf Neural Retrievers in Language Modeling}\label{sec:5.1}
We experimented with different off-the-shelf general purpose retrievers, and found that the sparse (lexical) BM25 retriever \cite{bm25} outperformed three popular dense (neural) retrievers: the self-supervised retrievers Contriever~\cite{izacard2022unsupervised} and Spider~\cite{ram-etal-2022-learning}, as well as a retriever based on the average pooling of BERT embeddings that was used in the RETRO system~\cite{retro}. We conducted a minimal hyper-parameter search on the query length $\ell$ for each of the retrievers, and found that $\ell=32$ was optimal for BM25 (Figure~\ref{fig:query_window}), and $\ell=64$ worked best for dense retrievers (Figures~\ref{fig:bert_query_window},~\ref{fig:contriever_query_window}).

Figure~\ref{fig:retrieval-comparison} compares the performance gains of In-Context RALM with these four general-purpose retrievers. The BM25 retriever clearly outperformed all dense retrievers. 
This outcome is consistent with prior work showing that BM25 outperforms neural retrievers across a wide array of tasks, when applied in zero-shot settings \cite{beir}. This result renders In-Context RALM even more appealing since applying a BM25 retriever is significantly cheaper than the neural alternatives.

\subsection{Frequent Retrieval Improves Language Modeling}\label{sec:5.2}

We investigated the effect of varying the retrieval stride $s$ (\textit{i.e.}, the number of tokens between consecutive retrieval operations).
Figure~\ref{fig:stride} shows that LM performance improved as the retrieval operation became more frequent. This supports the intuition that retrieved documents become more relevant the closer the retrieval query becomes to the generated tokens. Of course, each retrieval operation imposes a runtime cost. 
To balance performance and runtime, we used $s=4$ in our experiments. For comparison, RETRO employed a retrieval frequency of $s=64$~\cite{retro}, which leads to large degradation in perplexity. Intuitively, retrieving with high frequency (low retrieval stride) allows to ground the LM in higher resolution.

\subsection{A Contextualization vs. Recency Tradeoff in Query Length}\label{sec:5.3}
 
We also investigated the effect of varying $\ell$, the length of the retrieval query for BM25. 
Figure~\ref{fig:query_window} reveals an interesting tradeoff and a sweet spot around a query length of $32$ tokens. Similar experiments for dense retrievers are given in App.~\ref{app:query_length}. 
We conjecture that when the retriever query is too short, it does not include enough of the input context, decreasing the retrieved document's relevance. Conversely, excessively growing the retriever query deemphasizes the tokens at the very end of the prefix, diluting the query's relevance to the LM task.

\begin{figure}[t]
\centering
\hspace*{-10pt}
\includegraphics[width=1.05\columnwidth]{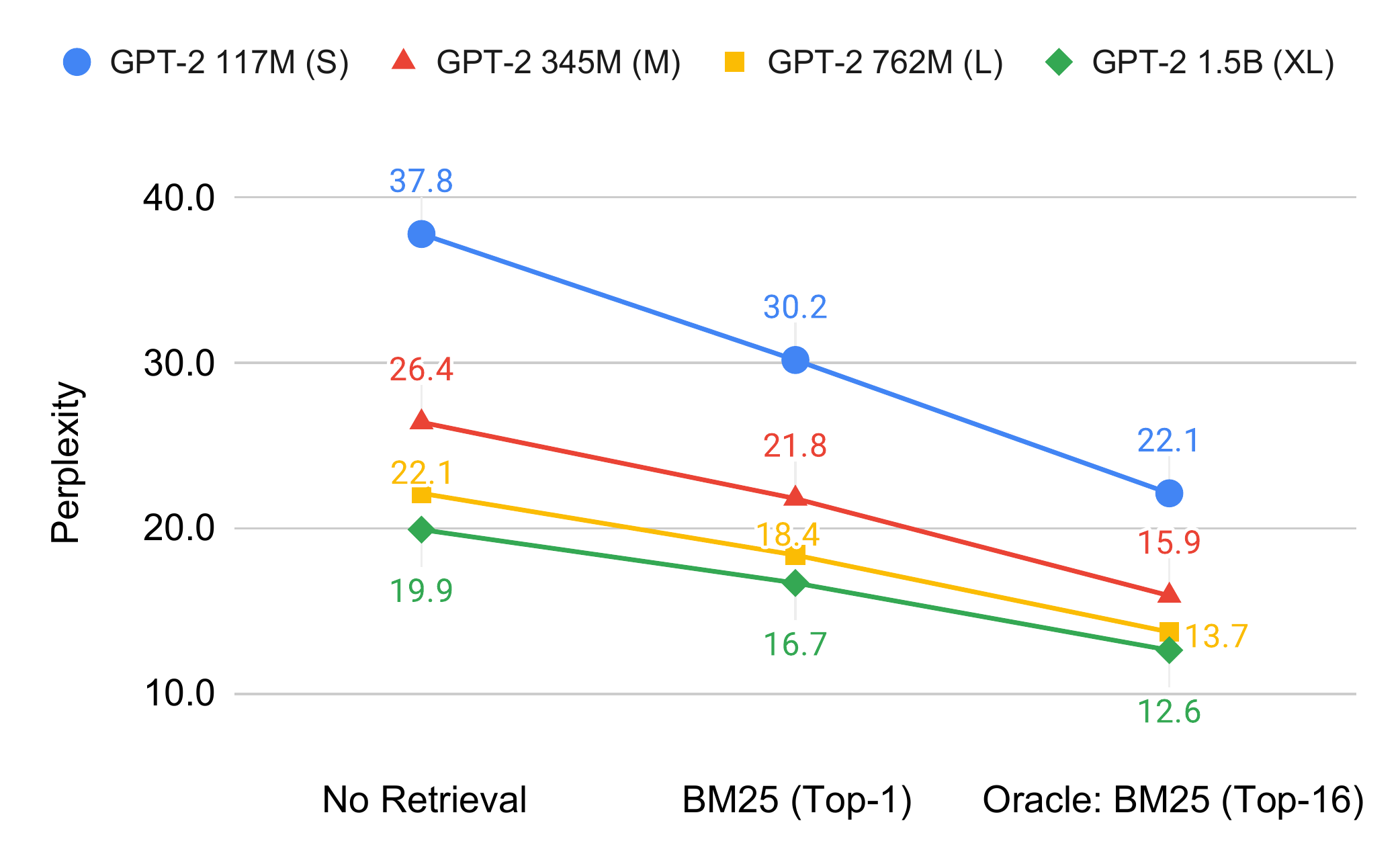}
\vspace{-15pt}
\caption{\textit{Potential for gains} from reranking: perplexity improvement (on the development set of WikiText-103) from an oracle that takes the best of the top-16 documents retrieved by BM25 rather than the first.
}
\label{fig:oracle}
\end{figure}
\section{Improving In-Context RALM with LM-Oriented Reranking}\label{sec:6}
Since In-Context RALM uses a fixed document reading component by definition, it is natural to ask whether performance can be improved by specializing its document retrieval mechanism to the LM task.
Indeed, there is considerable scope for improvement: the previous section considered conditioning the model only on the first document retrieved by the BM25 retriever.
This permits very limited semantic understanding of the query, since BM25 is based only on the bag of words signal. Moreover, it offers no way to accord different degrees of importance to different retrieval query tokens, such as recognizing that later query tokens are more relevant to the generated text. 

In this section, we focus on choosing which document to present to the model, by reranking the top-$k$ documents returned by the BM25 retriever.\footnote{In both \S\ref{sec:6.1} and \S\ref{sec:6.2} we use $k=16$.} 
We use Figure~\ref{fig:oracle} as motivation: it shows the large potential for improvement among the top-$16$ documents returned by the BM25 retriever. 
We act upon this motivation by using two rerankers.
Specifically, in \S\ref{sec:6.1} we show performance gains across our evaluation suite obtained by using an LM to perform zero-shot reranking of the top-$k$ BM25 retrieved documents (results in third row for each of the models in Table~\ref{tab:retrieval_results}). Then, in \S\ref{sec:6.2} we show that training a specialized bidirectional reranker of the top-$k$ BM25 retrieved documents in a self-supervised manner via the LM signal can provide further LM gains (results in forth row for each of the models in Table~\ref{tab:retrieval_results}).

\begin{table*}[t!]
\centering
\small
\begin{tabular}{llcc}
\toprule
\multirow{2.6}{0pt}{\textbf{Model}} & \multirow{2.6}{40pt}{\textbf{Reranking Model}} & 
\textbf{WikiText-103} & \textbf{RealNews}\\
\cmidrule(lr){3-3} \cmidrule(lr){4-4} 
& & word ppl & token ppl \\
\midrule
\multirow{2.1}{80pt}{\textbf{GPT-2 345M (M)} } &
GPT-2 110M (S) & 20.8 &  12.1  \\
& GPT-2 345M (M)  & 20.8 &  12.0 \\
\midrule
\multirow{2.1}{80pt}{\textbf{GPT-2 762M (L)}} &
GPT-2 110M (S) & 17.7 & 10.7  \\
& GPT-2 762M (L) & 17.6 & 10.6  \\
\midrule
\multirow{2.1}{80pt}{\textbf{GPT-2 1.5B (XL)} } &
GPT-2 110M (S) & 16.2 & 9.9 \\
& GPT-2 1.5B (XL) & 16.1 & 9.8 \\
\bottomrule
\end{tabular}
\caption{Perplexity for zero-shot reranking (\S\ref{sec:6.1}) where the reranking models is smaller than the LM, or the LM itself. 
Reranking is performed on the top 16 documents retrieved by BM25. 
Using a GPT-2 110M (S) instead of a larger language model as a reranker leads to only a minor degradation.}
\label{tab:table_gpt2_reranker}
\end{table*}

\subsection{LMs as Zero-Shot Rerankers}\label{sec:6.1}

First, we used off-the-shelf language models as document rerankers for the In-Context RALM setting. 
Formally, for a query $q$ consisting of the last $\ell$ tokens in the prefix of the LM input $x$, let $\{d_1,...,d_k\}$ be the top-$k$ documents returned by BM25. For retrieval iteration $j$, let the text for generation be $y:=x_{s\cdot j+1},...,x_{s\cdot j+s}$. Ideally, we would like to find the document $d_{i^*}$ 
that maximizes the probability of the text for generation, \textit{i.e.},
\begin{equation}\label{eq:future}
i^*=\arg\max_{i\in[k]} p_\theta(y|\left[d_i;x_{\leq s\cdot j}\right]). 
\end{equation}
However, at test time we do not have access to the tokens of $y$. 
Instead, we used the last \textit{prefix} tokens (which \textit{are} available at test time), denoted by $y'$, for reranking.
Formally, let $s'$ be a hyper-parameter that determines the number of the prefix tokens by which to rerank. 
We define $y':=x_{s\cdot j-s'+1},...,x_{s\cdot j}$ (\textit{i.e.}, the stride of length $s'$ that precedes $y$) and choose the document $d_{\hat{i}}$ such that
\begin{equation}\label{eq:zero-shot-reranking}
\begin{split}
    &\hat{i}=\arg\max_{i\in[k]} p_\phi(y'|\left[d_i;x_{\leq (s\cdot j-s')}\right]).
\end{split}
\end{equation}
The main motivation is that since BM25 is a lexical retriever, we want to incorporate a semantic signal induced by the LM. Also, this reranking shares conceptual similarities with the reranking framework of \citet{sachan-etal-2022-improving} for open-domain question answering, where $y'$ (\textit{i.e.}, the last prefix tokens) can be thought of as their ``question''. 

Note that our zero-shot reranking does not require that the LM used for reranking is the same model as the LM used for generation (\textit{i.e.}, the LM in Eq.~\eqref{eq:zero-shot-reranking}, parameterized by $\phi$, does not need to be the LM in Eq.~\eqref{eq:general_ralm}, parameterized by $\theta$). This observation unlocks the possibility of reranking with smaller (and thus faster) models, which is important for two main reasons: (i) Reranking $k$ documents requires $k$ forward passes; and (ii) it allows our methods to be used in cases where the actual LM's log probabilities are not available (for example, when the LM is accessed through an API).\footnote{Note we do not require that the two models share the same vocabulary.}

\paragraph{Results}
A minimal hyper-parameter search on the development set of WikiText-103 revealed that the optimal query length is $s'=16$,\footnote{We experimented with $s'\in\{4,8,16,32\}$.} so we proceed with this value going forward.
Table~\ref{tab:retrieval_results} shows the results of letting the LM perform zero-shot  reranking on the top-16 documents retrieved by BM25 (third row for each of the models). 
It is evident that reranking yielded consistently better results than simply taking the first result returned by the retriever.

Table~\ref{tab:table_gpt2_reranker} shows that a small LM (GPT-2 117M) can be used to rerank the documents for all larger GPT-2 models, with roughly the same performance as having each LM perform reranking for itself, supporting the applicability of this method for LMs that are only accessible via an API.

\subsection{Training LM-dedicated Rerankers}\label{sec:6.2}

Next, we \textit{trained} a reranker to choose one of the top-$k$ documents retrieved by BM25. 
We refer to this approach as \textit{Predictive Reranking}, since the reranker learns to choose which document will help in ``predicting'' the upcoming text.
For this process, we assume availability of training data from the target corpus. Our reranker is a classifier that gets a prefix $x_{\leq s\cdot j}$ and a document $d_i$ (for $i\in[k]$), and produces a scalar $f(x_{\leq s\cdot j}, d_i)$ that should resemble the relevance of $d_i$ for \textit{the continuation} of $x_{\leq s\cdot j}$. 

We then normalize these relevance scores:
\begin{equation}
    p_{\text{rank}}(d_i|x_{\leq s\cdot j})=\frac{\exp(f(x_{\leq s\cdot j}, d_i))}{\sum_{i'=1}^k \exp(f(x_{\leq s\cdot j}, d_{i'}))},
\end{equation}
and choose the document $d_{\hat{i}}$ such that
\begin{equation}
    \hat{i}=\arg\max_{i\in[k]} \ p_{\text{rank}}(d_i|x_{\leq s\cdot j}).
\end{equation}

\paragraph{Collecting Training Examples} To train our predictive reranker, we collected training examples as follows. 
Let $x_{\leq s\cdot j}$ be a prefix we sample from the training data, and $y:=x_{s\cdot j+1},...,x_{s\cdot j+s}$ be the text for generation upcoming in its next stride. We run BM25 on the query $q_j^{s,\ell}$ derived from $x_{\leq s\cdot j}$ (see \S\ref{sec:3.2}) and get $k$ documents $\{d_1,...,d_k\}$. For each document $d_i$, we then run the LM to compute $p_\theta(y|\left[d_i;x_{\leq s\cdot j}\right])$ similar to Eq.~\eqref{eq:ralm-hps}.

\begin{figure}[t]
\centering
\hspace*{-10pt}
\includegraphics[width=1.05\columnwidth]{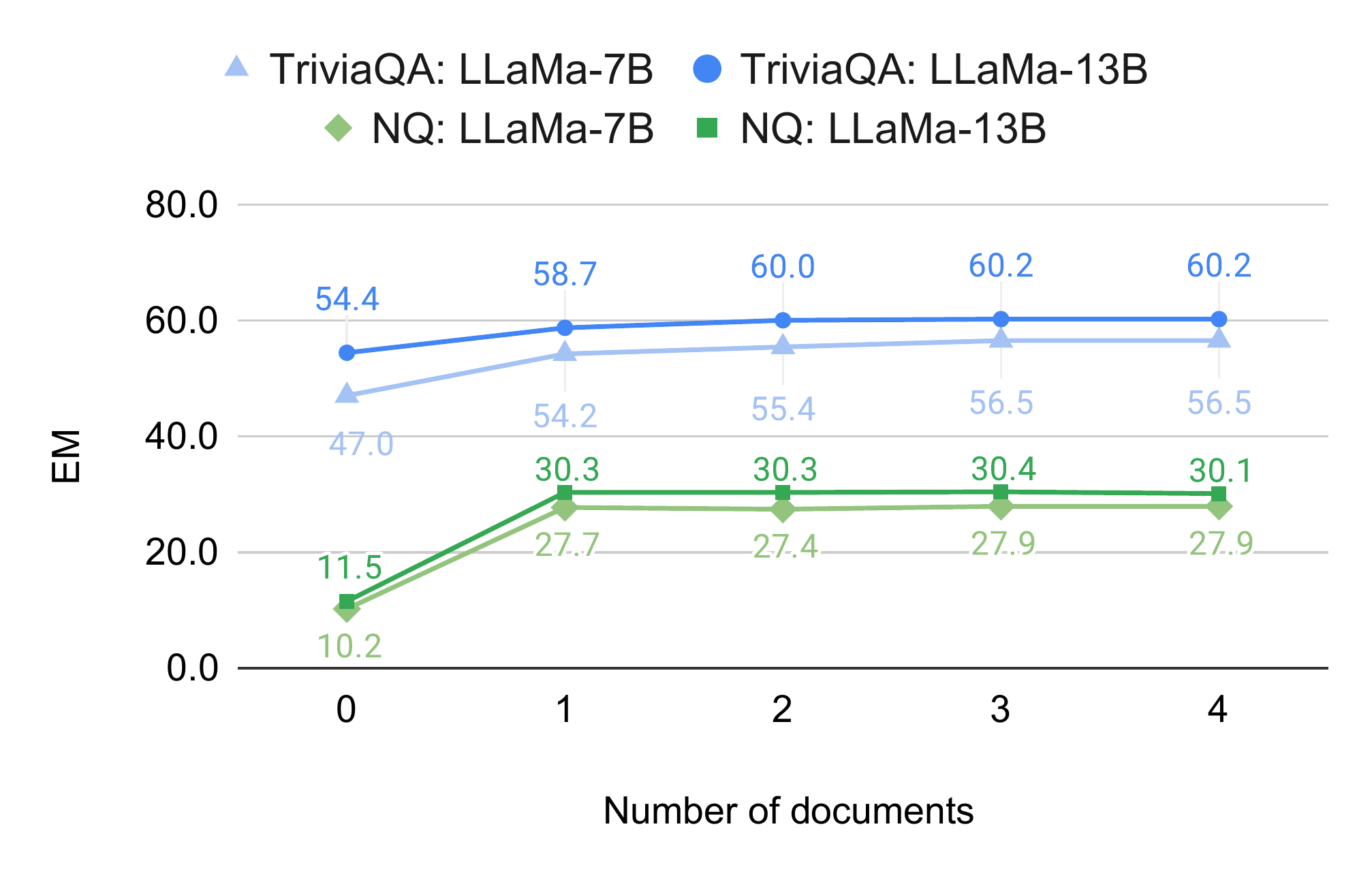}
\vspace{-15pt}
\caption{Zero-shot performance of In-Context RALM on the development set of Natural Questions and TriviaQA, when varying the number of documents (retrieved by DPR) shown in-context.}
\label{fig:odqa_ablations}
\end{figure}

\paragraph{Training} Our reranker was a fine-tuned RoBERTa-base \cite{liu2019roberta} that trained for 10,000 steps with a peak learning rate of $10^{-5}$ and a batch size of 32. Overall, we created 300,000 examples from the training set of WikiText-103 as explained above. 
The loss function we use to train the reranker follows previous work \cite{realm,rag}:
\begin{equation}
-\log\sum_{i=1}^k p_{\text{rank}}(d_i|x_{\leq s\cdot j})\cdot p_\theta(y|\left[ d_i;x_{\leq s\cdot j}\right]).
\end{equation}
Note that unlike those works, we train only the reranker ($p_{\text{rank}}$), keeping the LM weights $\theta$ frozen.

\paragraph{Results}
Table~\ref{tab:retrieval_results} shows the result of our predictive reranker, trained on WikiText-103. Specifically, we trained it with data produced by GPT-2 110M (S), and tested its effectiveness for all GPT-2 models. We observed significant gains obtained from Predictive Reranking. 
For example, the perplexity of GPT-2 110M (S) improved from 29.6 to 26.8, and that of GPT-2 1.5B (XL) improved from 16.6 to 15.4. This trend held for the other two models as well. Overall, these results demonstrate that training a reranker with domain-specific data was more effective than zero-shot reranking (Section~\ref{sec:6.1}).
Note that these results---while impressive---still leave room for further improvements, compared to the top-16 BM25 oracle results (see Figure~\ref{fig:oracle}). Moreover, the oracle results themselves can be improved by retrieving $k>16$ documents via a BM25 retriever, or by training stronger retrievers dedicated to the RALM task. We leave this direction for future work.

\begin{table}[t!]
\small
\centering
\begin{tabular}{llcc}
\toprule
\textbf{Model} & \textbf{Retrieval}~~~ & 
\textbf{NQ} & \textbf{TriviaQA}  \\
\midrule
\multirow{2.1}{60pt}{\textbf{LLaMA-7B} } &
- & 10.3 & 47.5 \\
& DPR  & 28.0 & 56.0 \\
\midrule
\multirow{2.1}{60pt}{\textbf{LLaMA-13B} } & - & 12.0 & 54.8  \\
& DPR & 31.0 & 60.1  \\
\midrule
\multirow{2.1}{60pt}{\textbf{LLaMA-33B} } & - & 13.7 & 58.3 \\
& DPR & 32.3 & 62.7 \\
\bottomrule
\end{tabular}
\caption{Zero-shot results of In-Context RALM on the test set of Natural Questions and TriviaQA measured by exact match. In the open-book setting, we include the top two documents returned by DPR.
}
\label{tab:odqa}
\end{table}
\section{In-Context RALM for Open-Domain Question Answering}\label{sec:7}

So far, we evaluated our framework on language modeling benchmarks. To test its efficacy in additional scenarios, and specifically downstream tasks, we now turn to evaluate In-Context RALM on open-domain question answering (ODQA;~\citealt{chen-etal-2017-reading}). 
This experiment is intended to verify, in a controlled environment, that LMs can leverage retrieved documents \textit{without further training} and \textit{without any training examples}.
Specifically, we use the LLaMA family \cite{touvron2023llama} \textit{with} and \textit{without} In-Context RALM (often referred to in ODQA literature as open-book and closed-book settings, respectively). In contrast to most prior work on ODQA (\textit{e.g.}, \citealt{izacard-grave-2021-leveraging,fajcik-etal-2021-r2-d2,izacard2022atlas,levine2022huge}), our ``reader'' (\textit{i.e.}, the model that gets the question along with its corresponding retrieved documents, and returns the answer) is simply a frozen large LM: \textit{not} pretrained, fine-tuned or prompted to be retrieval-augmented. 
For the closed-book setting, we utilize the prompt of \citet{touvron2023llama}.
For the open-book setting, we extend this prompt to include retrieved documents (see App.~\ref{app:odqa}).
We use DPR \cite{karpukhin-etal-2020-dense} as our retriever.

\paragraph{Varying the Number of Documents}
To investigate the the effect of the number of documents shown to the model, we performed a minimal analysis on the development set of NQ and TriviaQA.
Figure~\ref{fig:odqa_ablations} demonstrates that showing documents in-context significantly improves the model's performance. In addition, most of the gain can be obtained by using only two documents (or even a single one in some cases).

\paragraph{Results} Table~\ref{tab:odqa} gives the results of In-Context RALM on the test set of Natural Questions and TriviaQA. 
Motivated by our previous findings, we used two retrieved documents.
It is evident that showing the model relevant documents significantly boosted its performance. For example, adding retrieved documents improved LLaMA-13B in the zero-shot setting by more than 18 points on NQ (from 12.0\% to 31.0\%) and more than 5 points on TriviaQA (from 54.8\% to 60.1\%).
\section{Discussion}
Retrieval from external sources has become a common practice in knowledge-intensive tasks (such as factual question answering, fact checking, and more;~\citealt{petroni-etal-2021-kilt}). 
In parallel, recent breakthroughs in LM generation capabilities has led to LMs that can generate useful long texts. However, factual inaccuracies remain a common way in which machine-generated text can fall short, and lack of direct provenance makes it hard to trust machine generated text. This makes language modeling both a promising and an urgent new application area for knowledge grounding, and motivates promoting RALM approaches.  
Prior research has already investigated RALM, of course, but it is not yet widely deployed. One likely reason is that existing approaches rely upon fine-tuning the LM, which is typically difficult and costly, and is even impossible for LMs accessible only via an API.

This paper presented the framework of \textit{In-Context RALM}, enabling frozen, off-the-shelf LMs to benefit from retrieval. We demonstrated that substantial performance gains can be achieved by using general purpose retrievers, and showed that additional gains can be achieved by tailoring the document selection to the LM setting. A recent work by \citet{muhlgay2023generating} demonstrates that In-Context RALM is indeed able to improve the factuality of large LMs.

Several directions for further improvement remain for future work. First, this paper considers only the case of prepending a single external document to the context; adding more documents could drive further gains (for example,  using the framework of \citealt{ratner2022parallel}). Second, we retrieved documents every fixed interval of $s$ tokens, but see potential for large latency and cost gains by retrieving more sparsely, such as only when a specialized model predicts that retrieval is needed. 

We release the code used in this work, for the community to use and improve over. 
We hope it will drive further research of RALM, which will enable its wider adoption.

\section*{Acknowledgements}

We would like to thank the reviewers and the Action Editor for their valuable feedback.

\bibliography{anthology,custom}
\bibliographystyle{tacl_natbib}

\appendix

\section{Query Length Ablations}\label{app:query_length}

\begin{figure}[t]
\centering
\hspace*{-10pt}
\includegraphics[width=1.05\columnwidth]{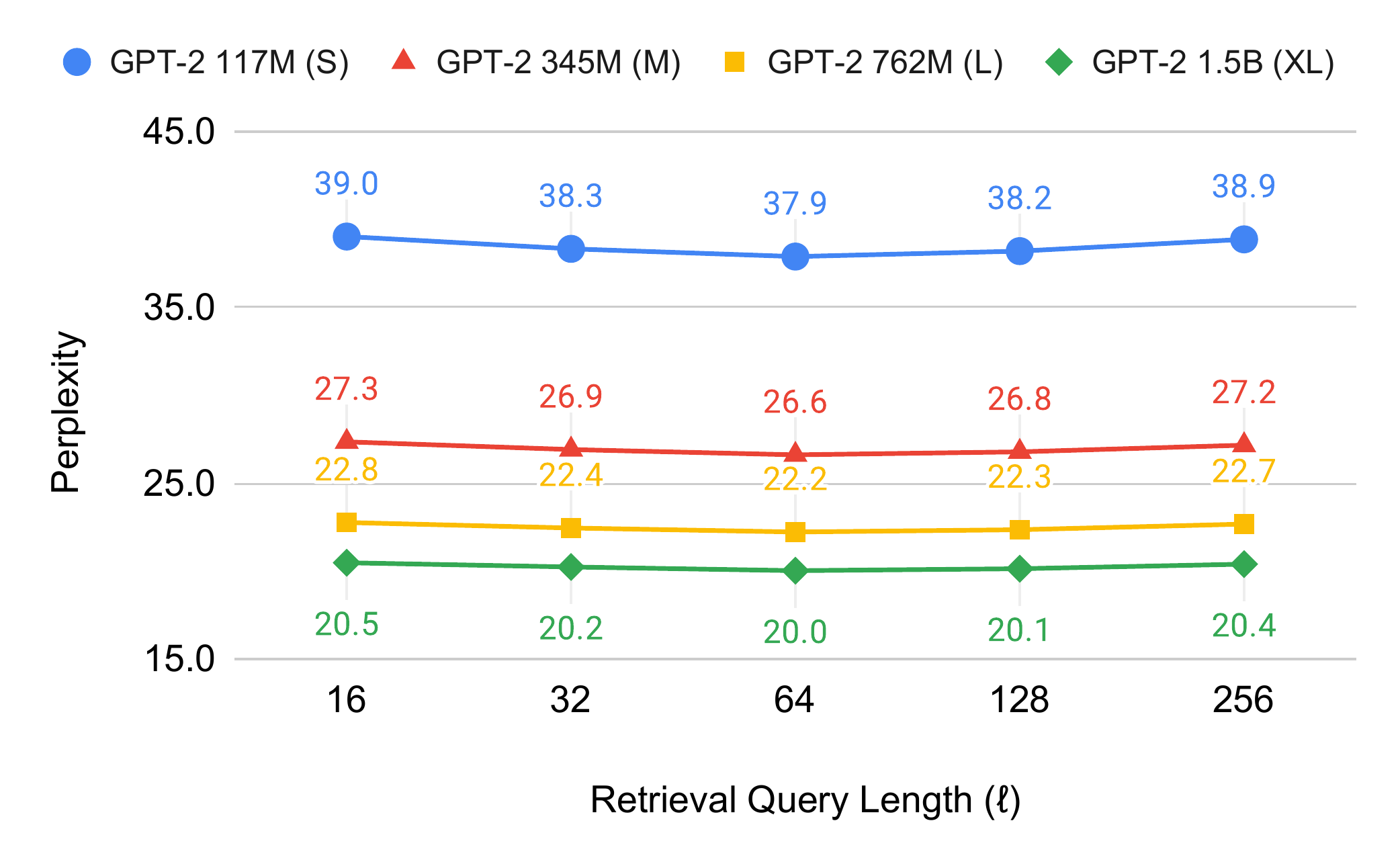}
\vspace{-15pt}
\caption{An analysis of perplexity as a function of \emph{the number of tokens in the query} for an off-the-shelf BERT retriever on the development set of WikiText-103. 
}
\label{fig:bert_query_window}
\end{figure}

\begin{figure}[t]
\centering
\hspace*{-10pt}
\includegraphics[width=1.05\columnwidth]{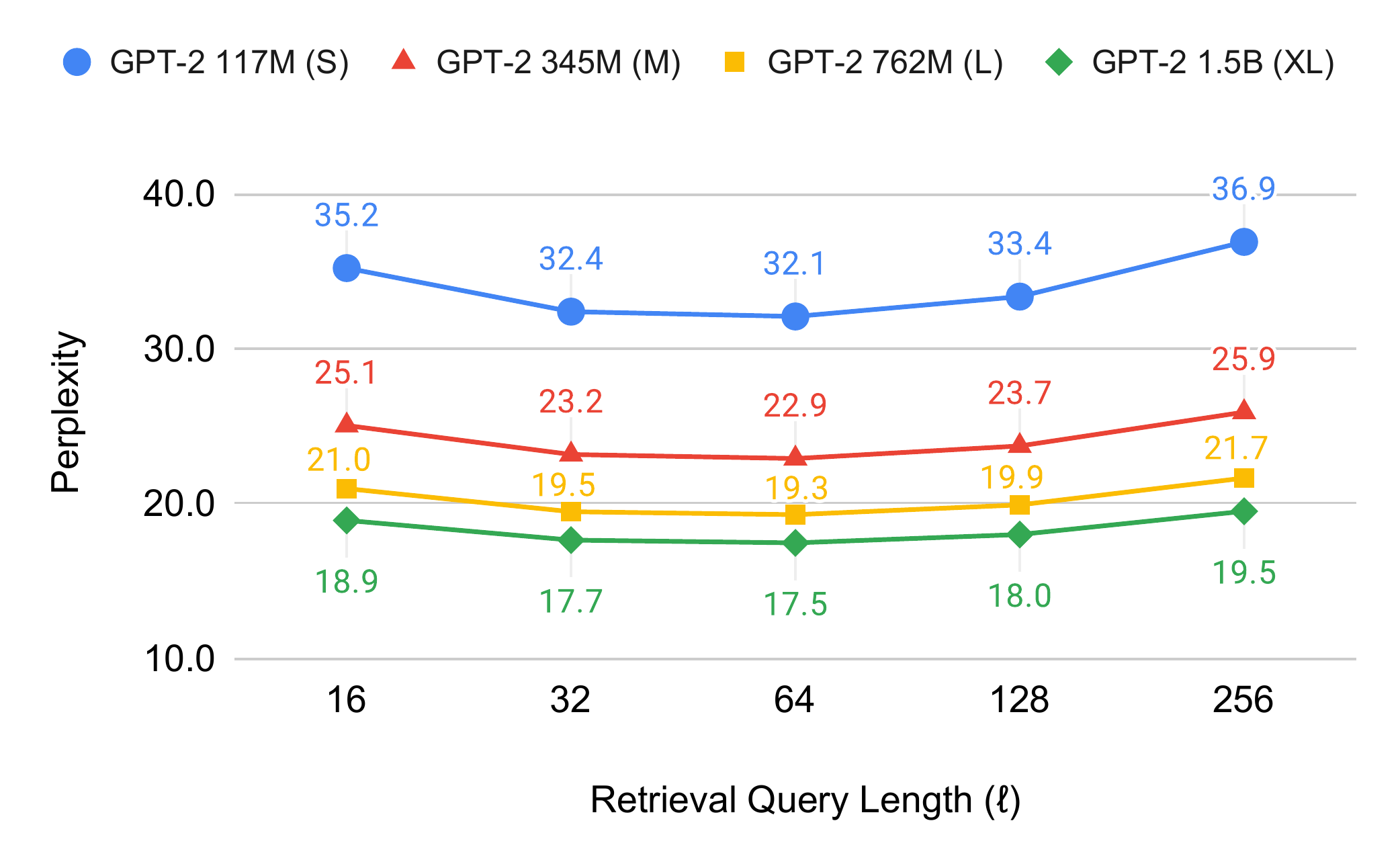}
\vspace{-15pt}
\caption{An analysis of perplexity as a function of \emph{the number of tokens in the query} for Contriever  on the development set of WikiText-103. 
}
\label{fig:contriever_query_window}
\end{figure}

Figure~\ref{fig:bert_query_window} and Figure~\ref{fig:contriever_query_window} show ablations on the optimal query length $\ell$ for off-the-shelf dense retrievers (BERT and Contriever respectively). We omit the results of Spider as they are almost identical to those of Contriever. Consistently, using $\ell=64$ (tokens) is optimal. This is in contrast to similar experiments we conducted for BM25 (\textit{cf}. Figure~\ref{fig:query_window}), where $\ell=32$ is optimal.

\section{GPT-Neo Results}\label{app:gpt-neo-results}
Table~\ref{tab:results-gpt-neo} gives the results of applying In-Context RALM to the models from the GPT-Neo model family on WikiText-103 and RealNews.

\section{Open-Domain Question Answering Experiments: Further Details}\label{app:odqa}

\paragraph{Closed-Book Setting} For the closed-book setting, we adopt the prompt of \citet{touvron2023llama}:
\begin{displayquote}
    \texttt{Answer these questions:}\\
    \texttt{Q: Who got the first nobel prize in physics?}\\
    \texttt{A:}
\end{displayquote}

\paragraph{Open-Book Setting} For the open-book setting, we extend the above prompt as follows:
\begin{displayquote}
    \texttt{Nobel Prize}\\ \\
    \texttt{A group including 42 Swedish writers, artists, and literary critics protested against this decision, having expected Leo Tolstoy to be awarded. Some, including Burton Feldman, have criticised this prize because they...}\\ \\
    \texttt{Nobel Prize in Physiology or Medicine}\\ \\
    \texttt{In the last half century there has been an increasing tendency for scientists to work as teams, resulting in controversial exclusions. Alfred Nobel was born on 21 October 1833 in Stockholm, Sweden, into a family of engineers...}\\ \\
    \texttt{Based on these texts, answer these questions:}\\
    \texttt{Q: Who got the first nobel prize in physics?}\\
    \texttt{A:}
\end{displayquote}

\begin{table}[t!]
\small
\centering
\begin{tabular}{@{}llcc@{}}
\toprule
\multirow{2.6}{0pt}{\textbf{Model}} & \multirow{2.6}{33pt}{\textbf{Retrieval}} & 
\textbf{Wiki-103} & \textbf{RealNews}   \\
\cmidrule(lr){3-3} \cmidrule(lr){4-4}
& & word ppl & token ppl \\
\midrule
\multirow{2.3}{63pt}{\textbf{GPT-Neo 1.3B} } &
- & 17.5 & 12.3 \\
& BM25, \S\ref{sec:5}  & 14.6 & ~~9.9 \\
\midrule
\multirow{2.3}{60pt}{\textbf{GPT-Neo 2.7B}} &
- & 15.1 & 11.0  \\
& BM25, \S\ref{sec:5} & 12.8 & ~~9.0  \\
\midrule
\multirow{2.3}{56pt}{\textbf{GPT-J 6B} } &
- & 11.6 & ~~9.2 \\
& BM25, \S\ref{sec:5}  & 10.0 & ~~7.7 \\
\bottomrule
\end{tabular}
\caption{The performance of models from the GPT-Neo family, measured by word-level perplexity on the test set of WikiText-103 and token-level perplexity on the development set of RealNews.
}
\label{tab:results-gpt-neo}
\end{table}


\end{document}